\documentclass{article}
\usepackage{amsmath}  
\usepackage{arxiv}
\usepackage{verbatim}
\usepackage[utf8]{inputenc} 
\usepackage[T1]{fontenc}    
\usepackage{hyperref}       
\usepackage{url}            
\usepackage{booktabs}       
\usepackage{amsfonts}       
\usepackage{nicefrac}       
\usepackage{microtype}      
\usepackage{cleveref}       
\usepackage{lipsum}         
\usepackage{graphicx}
\usepackage{natbib}
\usepackage{doi}
\usepackage{lmodern}
\usepackage{verbatim}
\usepackage{ulem}
\usepackage{xcolor}

\usepackage{bm}       
\usepackage{multirow}

\def\rset{\mathbb R}
\def\D{\mathcal{D}}
\def\F{\mathcal{F}}
\def\R{\mathcal{R}}
\def\by{\mathbf{y}}
\def\bx{\mathbf{x}}
\def\rmd{\mathrm{d}}
\DeclareMathSymbol{\dv}{\mathord}{operators}{"3A}

\title{Data-driven rainfall prediction at a regional scale: a case study with Ghana}

\newif\ifuniqueAffiliation

\ifuniqueAffiliation 
\author{ \href{https://orcid.org/0000-0002-2854-0709}{\includegraphics[scale=0.06]{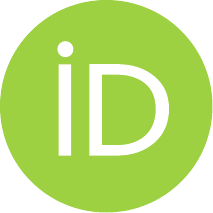}\hspace{1mm}Indrajit Kalita}\thanks{Corresponding autor} \\

}
\else
\usepackage{authblk}

\author[1]{%
	Indrajit Kalita\thanks{Corresponding author: \texttt{indrajit@bu.edu}}%
}
\author[2]{%
	Lucia Vilallonga%
}
\author[1,2]{%
	Yves Atchade%
}

\affil[1]{Faculty of Computing and Data Sciences, Boston University, Boston, USA}
\affil[2]{Department of Mathematics and Statistics, Boston University, Boston, USA}
\fi

\renewcommand{\headeright}{\textit{arXiv} preprint }
\renewcommand{\shorttitle}{Data-driven rainfall prediction}

\hypersetup{
pdftitle={Data-driven rainfall prediction at a regional scale: a case study with Ghana},
pdfsubject={q-bio.NC, q-bio.QM},
pdfauthor={Indrajit Kalita, Lucia Vilallonga, Yves Atchade},
pdfkeywords={Tropical Rainfall Prediction, ECMWF Forecast, Data-Driven forecasting, Convolutional Neural Ensemble learning, Feature Importance Estimation},
}

\begin{document}
\maketitle

\begin{abstract}
	With a warming planet, tropical regions are expected to experience the brunt of climate change, with more intense and more volatile rainfall events. Currently, state-of-the-art numerical weather prediction (NWP) models are known to struggle to produce skillful rainfall forecasts in tropical regions of Africa. There is thus a pressing need for improved rainfall forecasting in these regions. Over the last decade or so, the increased availability of large-scale meteorological datasets and the development of powerful machine learning models have opened up new opportunities for data-driven weather forecasting. Focusing on Ghana in this study, we use these tools to develop two U-Net convolutional neural network (CNN) models to predict 24h rainfall at 12h and 18h lead-time. The models were trained using data from the ERA5 reanalysis dataset and the GPM-IMERG dataset. A special attention was paid to interpretability. We developed a novel statistical methodology that allowed us to probe the importance of the meteorological variables input in our model, offering useful insights into the factors that drive precipitation in the Ghana region. Empirically, we found that our 12-hour lead-time model consistently delivers the strongest performance, and both our 12-hour and 18-hour lead-time models outperform the climatology forecast in every aspect. Additionally, their performance matches and sometimes surpasses the 18-hour lead-time forecasts produced by the ECMWF ensemble forecast (as available in the TIGGE dataset).
\end{abstract}


\section{Introduction}\label{sec:intro} 
The precipitation of liquid water from Earth's atmosphere to the surface is one of the most crucial atmospheric processes shaping life on Earth \citep{box2016}. Lack of rainfall leads to drought, and too much rainfall  leads to flooding, two extreme weather events that often lead to natural disasters. The spatio-temporal distribution and intensity of precipitation events also hold significant implications for the smooth running of human societies, from day-to-day events, to large-scale urban and agricultural plans. As such, precipitation prediction is an essential tool for planning and decision making. However, rain and other forms of precipitation are notoriously difficult to predict \citep{tapiador:etal:2019}. Unlike most atmospheric variables, rainfall amounts are spatio-temporally non-smooth, and this poses a significant challenge to modeling, with many models often based (implicitly or explicitly) on smoothness assumptions.

The situation is particularly problematic under the tropics where rainfall is dominated by multiscale convective flows that are currently challenging to represent in numerical weather prediction (NWP) models \citep{becker:etal:2021}. In tropical West Africa, this issue is compounded with the lack of quality measurements. As a consequence, state-of-the-art NWP models such as the European Center for Medium-range Weather Forecasts (ECMWF) models are known to struggle to produce skillful forecasts in these regions \citep{Vogel2020,Kniffka2020,Rojas2023}. At the same time, regions of sub-Saharan Africa are projected to experience the brunt of climate change, with more intense and more volatile precipitation as Earth's atmosphere warms \citep{africa:report:24}. For instance, flooding from the 2024 rainy season in West and Central Africa has claimed the lives of more than a thousand, and displaced close to a million people\footnote{\texttt{https://www.nytimes.com/2024/09/15/world/africa/floods-africa.html?smid=url-share}}. 

The rise of deep learning (DL) models and the growing availability of extensive weather data sets are ushering in a new set of tools and new opportunities for improved weather forecasting \citep{boualegue:etal:2024}. Following a similar trend in other areas, several foundation weather forecasting models have recently appeared with striking capabilities \citep{Pathak2022,Lam2022,bi2023accurate,Chen2023}. However, precipitation remains poorly handled by these foundation AI models and is often omitted from their evaluations. In some of these models, the approach adopted may not be well-suited for precipitation forecasting. For instance, the dynamical system perspective adopted in $\rm{GraphCast}$ uses a mesh-based graph convolutional neural network and is predicated on the smooth dynamics of the atmosphere \citep{Lam2022}, which typically does not apply to rainfall in the tropics. 

Beside foundation models, several other data-driven models have been developed for weather and precipitation prediction at a regional scale \citep{Shi2015,Dueben2018,Agrawal2019,Weyn2019,Ayzel2020,Ko2022,Oh2023}. Several works have also targeted specifically rainfall prediction in Sub-Saharan Africa using data-driven approaches \citep{Vogel2020,vogel:etal:2021,gebremichael:etal:2022,agee:etal:2023}, including \citet{walz:2024}, which uses deep learning techniques and is the closest to our work. Compared to \citet{walz:2024}, the prediction problem that we consider in this work is more challenging as we discuss below.

In this work, we tackle the tropical rainfall prediction problem at a smaller, regional scale. Unlike foundation models, we focus solely on rainfall prediction, and we build models that are more modest in size and can be easily deployed with limited resources. Specifically, using the 5th-generation Reanalysis (ERA5) of the ECMWF and the 6th-generation Integrated Multi-satellitE Retrievals for Global Precipitation Measurement (GPM-IMERG) datasets, we develop two deep learning (DL) models for 24h rainfall prediction over Ghana, with prediction lead-times of 12 hours and 18 hours, respectively. Overall, we found that our 12-hour lead-time model consistently delivers the strongest performance, and both our 12-hour and 18-hour lead-time models outperform the climatology forecast in every aspect. 
Additionally, we observed that using a hybrid approach that combines our data-driven model with classical NWP techniques further enhances forecast accuracy.

Our work focuses on an area $7.3^\circ$ by $5.6^\circ$ large, centered around Ghana at a 64-by-64 spatial resolution, compared to a 19-by-61 resolution in \citet{walz:2024} for an area 25 times bigger. The challenge in rainfall prediction increases significantly as the spatial resolution is refined. Furthermore, we consider predictions with longer lead-times (12h and 18h compared to 6h in \citet{walz:2024}). Another important contribution of our work is that we develop a statistical method to probe the predictive importance of the meteorological variables used in our model, leading to useful insights into the factors driving precipitation in the Ghana region.

The rest of the manuscript is structured as follows. Section \ref{sec:data} presents the general modeling approach and the datasets employed, and Section \ref{sec:method} discusses our methodology, specifically the U-Net architecture employed. Section \ref{sec:eval} describes the evaluation setup, including a novel approach for probing the importance of the input variables. In Section \ref{sec:res} we present and discuss our findings. Finally, Section \ref{sec:conclusion} contains some concluding thoughts.

\section{General approach and datasets}\label{sec:data}
\begin{figure}
	\centering
	\caption{Ghana: The area of interest within Africa.}
	\includegraphics[width = 0.8\textwidth, height=0.4\textheight]{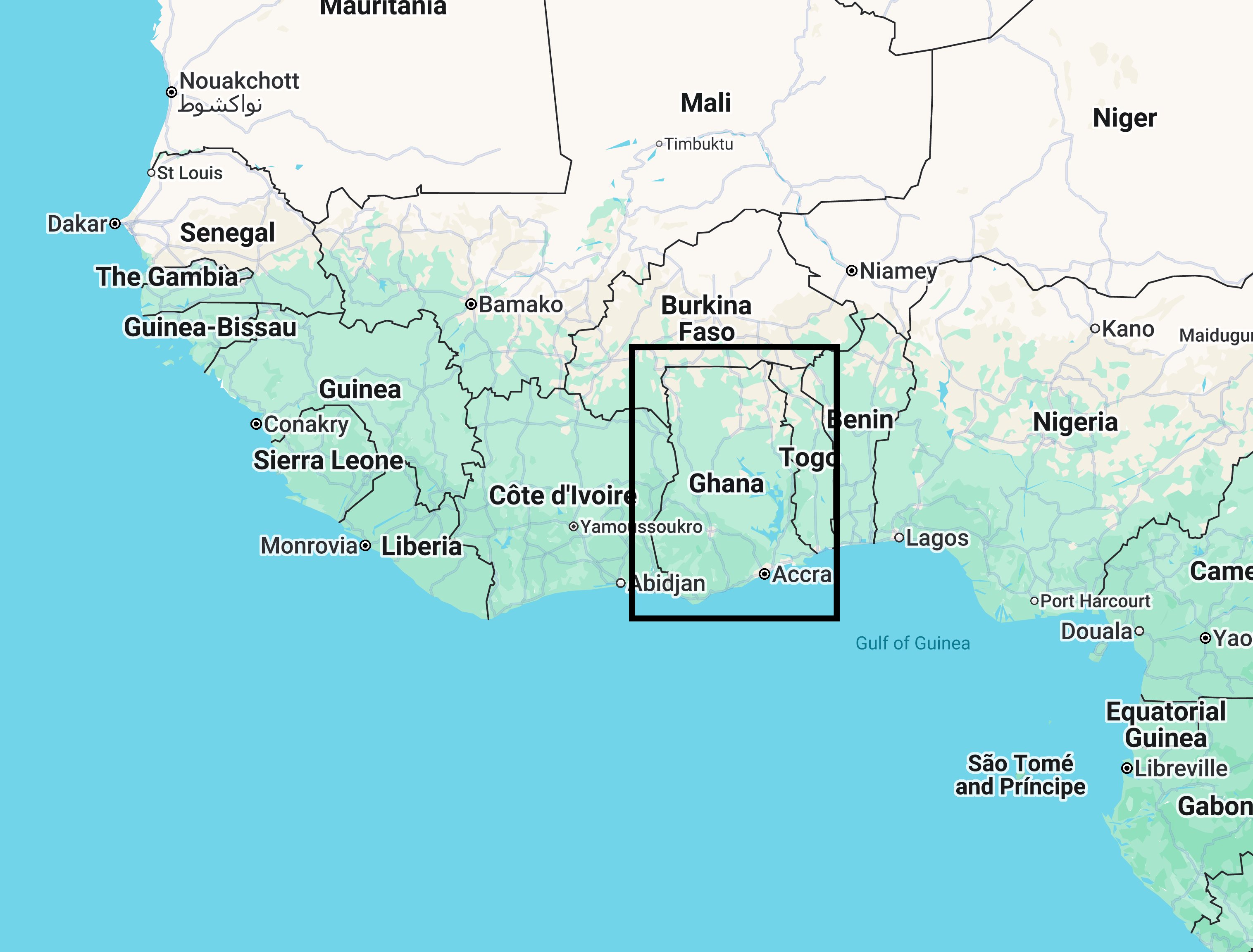}
	\label{AOI}\\
	{\footnotesize Source: Google. (n.d.). Partial map of Africa. Retrieved April 21st, 2025, using the Google map platform.}
\end{figure}

The current study focuses on an area centered around Ghana in West Africa, and spanning the latitudes $4.3^{\circ}$ North to $11.6^{\circ}$ North and the longitudes $3.8^{\circ}$ West to $1.8 ^{\circ}$ East, as depicted on Figure~\ref{AOI}. The climate of Ghana is tropical, with distinct rainy and dry seasons, with the rainy season usually lasting from April to November, followed by a dry season \citep{Manzanas2014}. The rainy season varies with the latitudes: the south typically experiences two rainy seasons -- a long rainy season from April to July, and a short one from September to November - whereas the north typically experiences one rainy season from April to October.

For this study, daily rainfall measurements and meteorological variables over Ghana were collected from the GPM-IMERG data product developed by the National Aeronautics and Space Administration (NASA), and from the ERA5 data product of the European Center for Medium-range Weather Forecasts (ECMWF). We describe these data products in Sections 2a-c below. Here we give a general overview of our approach. Specifically, the time-window of the study is June 1st, 2000, to September 30th, 2021, and we discretize the area of interest into a $64\times 64$ image. For each date $t$ in the study window, we let ${\bf y}_t = \{y_{i,t},\;i\in 1\dv 64\times 1\dv 64\}$ be the image of rainfall observations over the 24h of date $t$, obtained from the GPM-IMERG data product, where $1\dv 64$ is short for $\{1,\ldots,64\}$. We follow the convention of GPM-IMERG that defines the 24h rainfall period as 6AM-6AM UTC. Similarly, for each date $t$ in the study window, and for a prediction with lead-time $h$, we collected  $K=57$ environmental and meteorological variables over Ghana from the ERA5 data product at time $t-h$, as described in Section 2b below. We denote these variables on pixel $i$ by ${\bf x}_{i,t-h}\in\rset^K$, and we set ${\bf x}_{t-h} = \{{\bf x}_{i,t-h},\;i\in 1\dv 64\times 1\dv 64\}$ for the 3D images of meteorological predictor variables. 

For a prediction task with lead-time $h$, we thus obtain a dataset $\D =\{({\bf x}_{t-h},\;{\bf y}_t),\;1\leq t\leq T\}$ of size $T=7791$ that we split into a training set $\D^{'}$ and a test set $\D^{''}$ of size $7736$ ($100$ as validation samples) and $55$, respectively. Using the training set, we train a model that aims to predict ${\bf y}_t$ using the variables ${\bf x}_{t-h}$, assuming conditional independence between time points. We discuss the statistical assumptions in more details below in Section \ref{sec:method}. We explore two lead-time values: $h=12$, and $h=18$. And we evaluate and compare the models using the test set $\D^{''}$.

\subsection{GPM-IMERG precipitation data}
The Global Precipitation Measurement (GPM) mission is an international satellite mission launched jointly by NASA and the Japanese JAXA in 2014 for precipitation measurements at the scale of the planet \citep{Hou:etal:2014}. The mission was built on the successful Tropical Rainfall Measuring Mission (TRMM) satellites that operated from 1997 to 2015. The GPM mission carries the first space-borne Ku/Ka-band Dual-frequency Precipitation Radar (DPR) and a multi-channel GPM Microwave Imager (GMI). The Integrated Multi-satellite Retrievals for the Global Precipitation Measurement (IMERG) is a multi-satellite data processing algorithm developed by NASA \citep{Huffman2020}. The algorithm is applied to the data collected by TRMM and GPM, and further leverages whatever constellation of satellites is available at a given time to produce a long time-series of rainfall measurements at the scale of the planet \citep{Huffman2020}. The GPM-IMERG product is the current state-of-the-art satellite precipitation observation system, particularly under the tropics. The data is available at a 30-minute temporal resolution and 0.1-degree spatial resolution. For this study, we downloaded the 30-minute rainfall data and derived 24h accumulated (6AM-6AM UTC) GPM-IMERG version 6 (V6) data over the bounding box of Ghana at 0.1 degree spatial resolution, which we then regrid (cubic interpolation) to a $64\times 64$ image. The image obtained for date $t$ is the previously introduced ${\bf y}_t$. Although GPM-IMERG is known to have some bias \citep{gpm-imerg-eval:2022}, we will consider it as our precipitation ground truth.

\subsection{ERA5 meteorological variables} 
The ECMWF Reanalysis version 5 (ERA5) is the 5th generation of the gridded reanalysis climate and weather dataset maintained by ECMWF \citep{Hersbach2020}.  The dataset is produced through reanalysis, that is, through a data assimilation approach that combines numerical weather models with global climate observations at hourly temporal resolution. ERA5 data contains a large number of environmental and atmospheric variables,  from 1940 onwards, at an hour temporal resolution, and $0.25^{\circ}$ spatial resolution. The ERA5 dataset is driving much of the recent machine learning effort for data-driven weather forecasting \citep{Keisler2022,Lam2022,Chen2023}. We would like to note that the dataset is not available in real-time, as it currently has a latency of 5 days. As a result, our model, similar to most other data-driven weather prediction models that depend on ERA5 data inputs, is not yet operationally implementable.

Table \ref{tab:variables} lists the 55 ERA5 variables used for this project. We compiled this list from the literature, notably \cite{walz:2024}, as well as from a basic correlation analysis. We group the variables following \cite{walz:2024}. As noted in \cite{walz:2024}, all these variables have been reported in the literature as playing some role in rain formation processes. We retrieved these variables over the area of interest with a temporal resolution of six hours. 

\small
\begin{table*}[ht]
	\caption{ERA5 variables collected}
	\label{tab:variables}
	\centering
	\begin{tabular}{l|l|l}
		\hline
		\textbf{Variable} & \textbf{Name} & \textbf{Levels (in hPa)}\\
		\hline
		Specific rainwater content & crwc & 850, 925, 950 \\
		Total column water vapor & tcwv & single level \\
		Total cloud cover & tcc & single level \\
		Total column liquid water & tclw & single level \\
		Vertic. integ. moist. div. & vimd & single level \\\hline
		K index & kx & single level \\
		Convec. avail. pot. energy & cape & single level \\
		Convective inhibition & cin & single level \\\hline      
		Surface pressure & sp & single level \\
		Surface temperature & t2m & single level (2m) \\
		Dewpoint temperature & d2m & single level (2m) \\\hline
		Relative humidity & r & 300, 500, 600, 700, 850, 925, 950 \\
		Specific humidity & q & 300, 500, 600, 700, 850, 925, 950 \\
		Temperature & t & 300, 500, 600, 700, 850, 925, 950 \\
		Components of wind & u, v, w & 300, 500, 600, 700, 850, 925, 950 \\\hline
	\end{tabular}
\end{table*}
\normalsize

Since Ghana has a seasonal rainfall pattern that also varies strongly with latitude, accounting for space-time variability in the model is crucial for accurate forecasts. Thus in addition to the 55 variables listed in Table \ref{tab:variables}, for each pixel $i$ and each date $t$, we also constructed two new variables:
\begin{equation}\label{def:cos}
	\rm{COS}_{i,t} =\cos\left(2 \pi \frac{\rm{day}(t)}{365}\right) \times \rm{lat}(i),
\end{equation}
\begin{equation}\label{def:sin}
	\rm{SIN}_{i,t} =\sin\left(2 \pi \frac{\rm{day}(t)}{365}\right) \times \rm{lat}(i),
\end{equation}
where $\rm{day}(t)\in\{1,\ldots,365\}$ is the day of the year of date $t$, and $\rm{lat}(i)$ is the latitude of the center point of pixel $i$. 

In summary, for each date $t$ in our study window (between June 1st, 2000 and September 30th, 2021), and for each pixel $i\in 1\dv 64\times 1\dv 64$, we collected ${\bf x}_{i,t-h}\in\rset^{57}$, composed of the values of the 55 variables listed in Table \ref{tab:variables}, plus the two constructed variables $\rm{COS}$ and $\rm{SIN}$, $h$ hours before 6AM UTC of date $t$. It is worth mentioning that precipitation was not included as an input variable in the model. This allows our model to learn meaningful relationships between the available meteorological variables and rainfall, without reliance on a mechanistic auto-regressive model. 

\subsection{TIGGE forecast data}
We compare the skills of our models with the predictions from ECMWF, made available through the THORPEX Interactive Grand Global Ensemble (TIGGE) program \citep{swinbank:etal:2016}. The TIGGE database has established itself as a key resource for evaluating the capabilities of state-of-the-art operational NWP models and has proven invaluable to the research community. For this project, we use the Noon forecasts from the ECMWF ensemble (50 perturbed members + 1 control member) available through TIGGE. The ECMWF noon-forecast predicts the state of the atmosphere every 6h starting from Noon. From these forecasts, we derived the predicted 24h accumulation of rainfall from 6AM (next day) to 6AM the day after next, corresponding to an 18h lead-time prediction. We focus on 6AM-6AM to match the GPM-IMERG precipitation window. We downloaded the total rainfall (the sum of large-scale organized precipitation and convective precipitation) from the ECMWF ensemble forecasts and re-gridded it to $64\times 64$ images for further analysis.

\section{Methodology}\label{sec:method}
We relate the rainfall image observation ${\bf y}_t$ to the environmental and meteorological variables ${\bf x}_{t-h}$ via the model
\begin{equation}\label{eq:main:model}
{\bf y}_t = \mathcal{F}_W\left({\bf x}_{t-h}\right) + \bm{\epsilon}_t,\;\;\; t=1,\ldots,T,
\end{equation}
for some error term $\bm{\epsilon}_t$, and for a regression function $\mathcal{F}_W$ that we build as a U-Net function with weight parameter $W$ \citep{Ronneberger2015}. The parameter $W$ represents the set of all trainable variables of the U-Net function. We refer the reader to \citep{zhang2023dive} for an introduction to deep learning. We assume that conditionally on $\{{\bf x}_{t-h},\;1\leq t\leq T\}$ the error terms $\{\bm{\epsilon}_{i,t},\;1\leq t\leq T,\;i\in 1\dv 64\times 1\dv 64\}$ are independent mean-zero and identically distributed. This is clearly a strong assumption, and it may be possible to improve the model by further leveraging space-time dependence in the error terms. We leave this issue for further research. We note however that space-time dependence is captured in the conditional mean of ${\bf y}_t$ given ${\bf x}_{t-h}$ through the input variables {\rm COS} and {\rm SIN} and through the convolutional layers of the function $\mathcal{F}_W$. 

\begin{figure}
\caption{The U-Net architecture employed.  In each block, the first dimension indicates the number of feature maps, while the second and third dimensions correspond to the width and height of each feature map. The input dimensions consist of 57 (representing weather variables), followed by a width of 64 and a height of 64, respectively.}
\includegraphics[width=\textwidth]{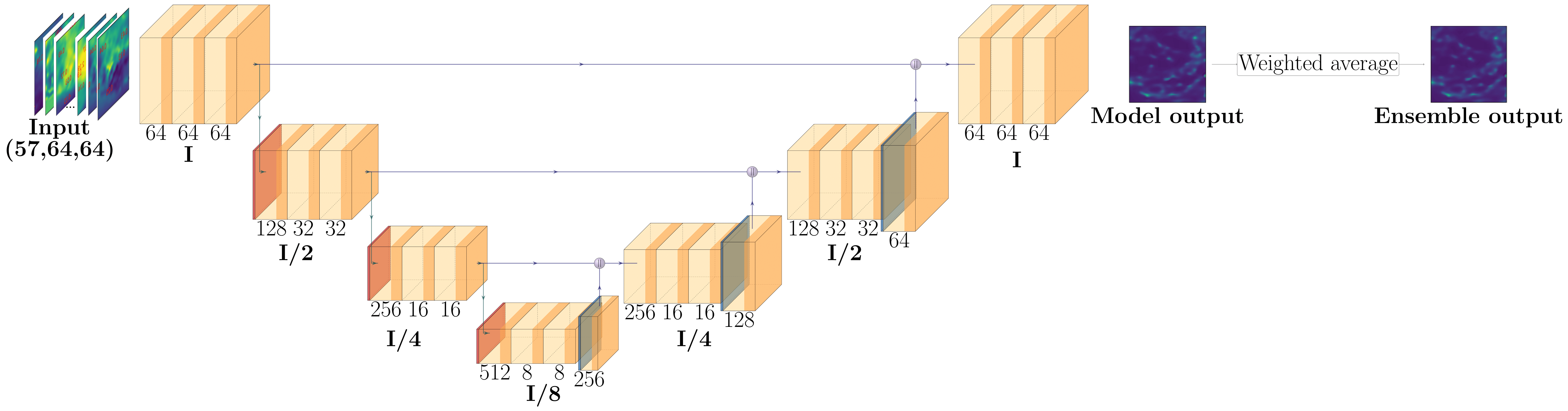}
\label{fig:unet}
\end{figure}

The U-Net architecture is a CNN-based deep neural network (DNN) architecture specifically designed for image tasks. Specifically, we utilize a 2D U-Net model with three downsampling blocks, each containing two convolutional layers, followed by three upsampling blocks with concatenation skip connections (see Figure \ref{fig:unet}). In contrast to more computationally expensive architectures like graph neural networks (GNNs) and Transformers, U-Net models offer a balance between accuracy and efficiency, and have been used for weather data in several prior works \citep{Lam2022,Keisler2022,walz:2024}. The model leverages an encoder-decoder structure with concatenation skip connections to achieve high segmentation accuracy while maintaining computational efficiency. The encoder layers map the input data to a more abstract/latent space. After further abstraction through the convolutions in the middle layers, the decoder layers map the data back to a gridded image. The skip connections concatenate same-level encoding and decoding layers, which is believed to improve performance for image-to-image problems \citep{Ronneberger2015}, although a rigorous understanding remains an open problem.

Using the same U-Net architecture, we trained two models with different lead-times. The first model has a lead-time $h=12$ hour that we denote by \rm{UNET$_{12}$} (this model uses as input, the state of the meteorological variable at 6 PM, 12 hours before the 6AM-6AM rainfall window). The second model has a lead-time of $h=18$ hours, and is denoted \rm{UNET$_{18}$}  (based on meteorological variables at Noon, 18 hours before the 6AM-6AM rainfall window).

Given the training data $\D'$, we train the U-Net models (\ref{eq:main:model}) by minimizing the L1 loss in (\ref{eq:loss}) with respect to $W$:
\begin{equation}\label{eq:loss}
\rm{L}(W;\D') =  \sum_{t\in\D'} \; \left\| \textbf{y}_{t} - \mathcal{F}_W\left({\bf x}_{t-h}\right) \right\|_1,
\end{equation}

where for an image ${\bf a}$, $\|{\bf a}\|_1$ denotes the sum of the absolute values of the pixel values of ${\bf a}$. The L1 loss is used for added robustness to outliers.  We solve this minimization problem using the Adam optimizer \citep{adam:2014}  with a learning rate of $10^{-4}$ and a weight decay of $10^{-4}$. During training, a batch size of $256$ is used (based on the availability of the GPU memory). We adjust the learning rate dynamically during training, using the \texttt{ReduceLROnPlateau} method \cite{Krizhevsky2012}. This scheduler monitors the loss and adjusts the learning rate when a plateau in performance is detected. Specifically, the learning rate is reduced by a factor of $0.1$  if no improvement is observed in the loss for a specified patience period of $100$ epochs.

If $\widehat{W}$ is an approximate solution to the minimization of  (\ref{eq:loss}), and given a new test data point ${\bf x}_{t'-h}\in\D^{''}$, we predict the rainfall amount at time $t'$ using
\[\hat {\bf y}_{t'} = \mathcal{F}_{\widehat{W}}\left({\bf x}_{t'-h}\right).\]
For each test date $t'\in\D^{''}$, the prediction $\hat {\bf y}_{t'}= \{\hat y_{i,t'},\;i\in 1\dv 64\times 1\dv 64\}$ is a $64 \times 64$ image, where $\hat y_{i,t}$ is the prediction at pixel $i$. We evaluate the performance of the model by comparing $\hat {\bf y}_{t'}$ and ${\bf y}_{t'}$ over the test set $\D^{''}$, as we describe in the next section. 

\subsection{Uncertainty quantification using EasyUQ} \label{sec: easyuq}
For uncertainty quantification, and for a meaningful comparison with the TIGGE ensemble predictions, we use the EasyUQ method developed by \cite{walz:etal:2024a} to estimate the cumulative distribution function (cdf) of the conditional distribution of the ground truth $y_{i,t}$ given $\hat{y}_{i,t}$. This provides a measure of uncertainty on the prediction $\hat{y}_{i,t}$. 

These estimated CDFs are obtained by solving an isotonic regression problem on the training dataset, with interpolation adjustments on the test dataset. We give a brief description of the method, and refer the reader to \citet{henzi:etal:2021,walz:etal:2024a} for more details, and for software. Fix a pixel $i$, and let $\{(y_{i,t},\hat{y}_{i,t}),\;t\in\mathcal{D}^{'}\}$ be the ground truths and prediction values in the training set at pixel $i$. Let $F_{\hat{y}_{i,t}}$ denote the conditional cdf of $y_{i,t}$ given $\hat{y}_{i,t}$. EasyUQ estimates  the map $\{\hat{y}_{i,t},\;t\in\mathcal{D}^{'}\}\mapsto \{F_{\hat{y}_{i,t}},\;t\in\mathcal{D}^{'}\}$ (a distributional regression problem) by minimizing the loss
\[ \min_{\{F_{i,t},\;t\in\mathcal{D}^{'}\}}\;\sum_{t\in\mathcal{D}^{'}} {\rm CRPS}(F_{i,t},y_{i,t}),\;\;\;\; \mbox{ where }\;\;\;\;  {\rm CRPS}(F,y) = \int_{-\infty}^{+\infty}\left(F(u) - \textbf{1}_{\{y\leq u\}}\right)^2\rmd u,\]
is the continuous ranked proper scoring (CRPS) of cdf $F$ and target $y$ (see \cite{gneiting:raftery:2007} for more details on the CRPS). The minimization problem described above is a minimization over cdfs $\{F_{i,t},\; t\in\mathcal{D}^{'}\}$. Under a monotonicity assumption, it was shown by \cite{henzi:etal:2021} that this minimization problem has a unique solution $\{\hat F_{i,t},\;t\in\mathcal{D}^{'}\}$ that can be easily computed. $\hat F_{i,t}$ is the EasyUQ estimate of $F_{\hat{y}_{i,t}}$. When apply on the sample $t'$ in the test set $\mathcal{D}^{''}$, $F_{\hat y_{i,t'}}$ is estimated by interpolation from the estimates $\{\hat F_{i,t},\;t\in\mathcal{D}^{'}\}$ in the training set.

\subsection{Combining \rm{UNET$_{18}$} and ECMWF ensemble forecasts\label{sec:hyb}}

In addition to the two models \rm{UNET$_{12}$} and \rm{UNET$_{18}$}  we also explore a prediction model that combines \rm{UNET$_{18}$} with the 18h lead-time forecast of the ensemble ECMWF (referred as NWP forecast) obtained from the TIGGE database. This hybrid approach takes stock of the distinctive features of both methodologies, potentially yielding a more accurate forecast. For $m\in\{{\rm UNET}_{18},\; {\rm NWP}\}$, we write $\hat y_{i,t}(m)$ to denote the prediction of method $m$ for time $t$ and pixel $i$. We define the hybrid prediction as

\begin{equation}\label{eq:hyb}
\hat{y}_{i,t}(\mathrm{HYB}) = (1 - \beta) \cdot \hat{y}_{i,t}(\mathrm{UNET}_{18}) + \beta \cdot \hat{y}_{i,t}(\mathrm{NWP}),
\end{equation}
where the  $\beta$ is estimated via linear regression over the test set by regressing the true residual \( y_{i,t} - \hat{y}_{i,t}(\mathrm{UNET}_{18}) \) on the difference \( \hat{y}_{i,t}(\mathrm{NWP}) - \hat{y}_{i,t}(\mathrm{UNET}_{18}) \), without an intercept. 
We obtain \( \beta = 0.546 \), indicating that the hybrid is almost an equal-weight average of both models, with slightly more weight given to the NWP forecast.

We obtain an ensemble of hybrid predictions by applying Equation (\ref{eq:hyb}) to each of the 51 members of the {\rm NWP} ensemble. Note that the EasyUQ method of \cite{walz:etal:2024a} cannot be applied to the hybrid prediction here, since we do not have the TIGGE predictions on the training dataset.
%

\section{Model evaluation}\label{sec:eval}
We evaluate and compare four models: \rm{UNET${_{12}}$}, \rm{UNET${_{18}}$}, the 18-hour lead-time predictions of ECMWF, obtained from TIGGE and denoted as \rm{NWP}, and the hybrid model \rm{HYB}, which is a combined version of \rm{UNET$_{18}$} and \rm{NWP}.  We use the climatology prediction as reference, and denote it as \rm{CLIM}. The climatology-based rainfall prediction at location $i$ and date $t$ is defined as the average of all the 24h rainfall values from the GPM-IMERG dataset recorded in previous years at the same location $i$, considering the same calendar day $t$, as well as two days before and two days after. We also use this ensemble of past rainfall values as an empirical CDF for the purpose of uncertainty quantification. We compare these models using the mean absolute error (MAE), the CRPS as defined above, as well as the precision, recall, and F1-score for rainfall event detection.

\subsection{Prediction error in terms of MAE and CRPS}
For a given model $m\in\{ \rm{CLIM, UNET_{12}, UNET_{18}, NWP, HYB}\}$, we write $\hat y_{it'}(m)$ to denote the predicted rainfall at location $i\in 1\dv 64\times 1\dv 64$, and at time $t'$ by model $m$. The mean absolute prediction error (MAE) of model $m$ at pixel $i$ is defined as
\[\ell(i,m) =\frac{1}{|\D^{''}|}\sum_{t'\in\D^{''}} \left|\widehat{y}_{i,t'}(m) - y_{i,t'}\right|,\]
where the average is taken over the test dataset $\D^{''}$. Similarly, let $\hat{F}_{i,t'}^{(m)}(\cdot)$ denote the estimated cdf for method $m$ at time $t'$ and pixel $i$. The CRPS of model $m$ at pixel $i$ is defined as:
\[{\rm CRPS}(i,m) = \frac{1}{|\D^{''}|}\sum_{t'\in\D^{''}} {\rm CRPS}(\hat{F}_{i,t'}^{(m)}, y_{i,t'}).\]
In our setting, all the cdfs $\hat{F}_{i,t'}^{(m)}(\cdot)$ are discrete, and the CRPS can be efficiently computed as in \citep{jordan:etal:2019}. The CRPS reduces to the MAE for ensembles with a single element. Using the CLIM prediction as a reference, we compute the pixel-by-pixel CRPS error skill of model $m$ as

\begin{equation}\label{eq:skill}
\rm{Skill}(i,m) = \frac{ {\rm CRPS}(i,{\rm CLIM}) - {\rm CRPS}(i,m)}{{\rm CRPS}(i,{\rm CLIM})},
\end{equation}

which offers a visual evaluation of model $m$. As defined, $\rm{Skill}(i,m)$ is positive if model $m$ performs better than the CLIM forecast at $i$, and $\rm{Skill}(i,m)$ is nonpositive otherwise.

\subsection{Precision, recall, and F1-score}
We also compare the models in their ability to correctly predict rainfall events. Given a threshold level $\tau$, and a model $m\in\{ \rm{CLIM, UNET_{12}, UNET_{18}, NWP, HYB}\}$, its precision and recall at pixel $i$ are defined respectively as
\begin{equation*}
\mathcal{P}_i(m) = \frac{\sum_{t'\in\D^{''}}\textbf{1}_{\left\{|\hat y_{it'}|>\tau\right\}}\textbf{1}_{\left\{|y_{it'}|>\tau\right\}}}{\sum_{t'\in\D^{''}}\textbf{1}_{\{|\hat y_{it'}|>\tau\}}},\;\; \R_i(m) = \frac{\sum_{t'\in\D^{''}}\textbf{1}_{\{|\hat y_{it'}|>\tau\}}\textbf{1}_{\{|y_{it'}|>\tau\}}}{\sum_{t'\in\D^{''}}\textbf{1}_{\{|y_{it'}|>\tau\}}}.
\end{equation*}
The recall statistic $\R_i(m)$ measures the proportion of days in the test set with rain amount larger than $\tau$ correctly detected by  method $m$ (at location $i$). The precision $\mathcal{P}_i(m)$ is the true positive rate of method $m$ at location $i$: among the days that method $m$ assigned as having rain amount larger than $\tau$, the proportion of them that indeed have rain amount larger than $\tau$. We further combine the precision and the recall into a  $(\mathcal{F}_1)$ score
\[ (\F_1)_{i}(m) = \frac{2 \mathcal{P}_i(m) \mathcal{R}_i(m)}{\mathcal{P}_i(m) + \mathcal{R}_i(m)}.\]
We further average the precision statistic $\mathcal{P}_i(m)$ across pixels to obtain the overall precision of model $m$ as
\[\mathcal{P}(m) = \frac{1}{4096}\sum_{i} \mathcal{P}_i(m).\]
We compute $\mathcal{R}(m)$ and $(\mathcal{F}_1)(m)$ similarly. We compute these statistics with two threshold values: $\tau=0.5$ mm, which allows us to compare the models in their ability to predict rainfall events, and $\tau=10$ mm to evaluate the ability of the models to predict heavy rainfall events. For the threshold $\tau = 0.5$ mm, 34.9\% of all grid cells have rainfall above this threshold, while 46.1\% of rainy grid cells (i.e., rainfall greater than 0\,mm) exceed it. At $\tau = 10$ mm, the corresponding values are 10.0\% and 13.2\%, respectively. Overall, 75.6\% of grid cells in the dataset record nonzero rainfall.

\subsection{Probing the importance of the input variables}\label{sec:varsel}
One of the main limitations of deep learning models is their lack of interpretability. With the development and adoption of these techniques in Earth science research, there is a growing need for interpretable/explainable deep learning models \citep{bommer:etal:2024}. 
We focus here on our model $\rm{UNET}_{12}$, and we propose a novel methodology to evaluate the importance of its input predictors.  Let $\F_{\hat W}$ denote the fitted model $\rm{UNET}_{12}$. We recall that the input feature ${\bf x}$ is of dimension ${\bf x}\in\rset^{K\times I}$, where $K$ denote the number of variables, and $I$ the number of pixels (we have collapsed the longitude and latitude dimensions into one pixel dimension). A common approach that one could use to measure the importance of a given input feature in this setting is the sensitivity of the predicted rainfall with respect to that input \citep{baehrens:etal:2010,bommer:etal:2024}. Specifically, the sensitivity of the predicted rainfall to the input variable $k$ is defined as
\[\frac{1}{|\D^{''}|} \sum_{t'\in\D^{''}}\;\sum_{\iota}\;\left|{\bf s}_{k,\iota}({\bf x}_{t'})\right|,\;\;\mbox{ where }\;\; {\bf s}_{k,\iota}({\bf x}) = \frac{1}{2} \frac{\partial\|\F_{\hat W}({\bf x})\|_2^2}{\partial {\bf x}_{k,\iota}},
\]
where the average is over the test dataset $\D^{''}$. \citet{bommer:etal:2024} presented several variations of this metric that have been used in the literature. 
We note however, that these sensitivity metrics are defined without any reference to model performance. Hence, a variable could have high sensitivity without being essential for accurate prediction. 

To address this issue, we propose a novel methodology built on the idea of posterior predictive checking, a powerful tool in Bayesian data analysis for model checking and validation \citep{gelman:etal:2004}. The basic idea is the use of the posterior predictive distribution as a data-generating process for the test data. Based on that data-generating process, we then perform a variable selection to determine which input variable best explains the observed performance on the test data. To be more specific, suppose for a moment that we have fitted our deep neural network model in the Bayesian framework with a posterior distribution $\Pi(\cdot\vert \D')$. Then the posterior predictive distribution of a data point $\bx_{t-h}$ in the test set $\D^{''}$ would be defined as
\[\hat f(\bx_{t-h}) = \int \exp\left(-\frac{1}{2\sigma^2}\| \by_t - \F_W(\bx_{t-h})\|_1\right)\Pi(W\vert\D') \rmd W.\]
To the $K=57$ input variables in model (\ref{eq:main:model}), we assign binary variables $\delta=(\delta_1,\ldots,\delta_K)$, where $\delta_k\in\{0,1\}$. Let $\delta\odot\bx_t$ be the input data point with the same size as $\bx_t$, but where the channel of the $j$-th variable is multiplied by $\delta_j$. In other words, with $\delta\odot\bx_t$, we mask out from the input data, all the variables $j$ for which $\delta_j=0$. We endow each $\delta_j$ with a Bernoulli prior distribution: $\delta_j\sim \textbf{Ber}(1/(1+e^r))$, for some hyper-parameter $r$ to be set by the user. Note that under this prior distribution, the probability $\mathbb{P}(\delta_k=1)$ is of order $e^{-r}$, and thus can be very small even for $r$ modestly large. Given this prior distribution and the posterior predictive model, we consider the posterior distribution (from the posterior predictive model)
\begin{equation}\label{post:dist}
\Pi(\delta_1,\ldots,\delta_K\vert \D'\cup\D^{''})\propto \exp\left(-r\sum_{j=1}^K \delta_j \right)\;\prod_{t\in\D^{''}}\; \hat f(\delta\odot\bx_{t-h}),\end{equation}
which gives the relative importance of each variable $\delta_j$ in explaining the performance of our fitted model $\Pi(W\vert\D')$ on the test data $\D^{''}$.  We note that the posterior predictive distribution $\Pi(\delta\vert \D'\cup\D^{''})$ in (\ref{post:dist}) depends both on the test set $\D{''}$ and the training set $\D'$. We note also that the test set $\D^{''}$ is typically not very large; therefore, we do not expect $\Pi(\delta\vert \D'\cup\D^{''})$  to concentrate, in the sense of putting high probability on a small number of variables. Instead, it provides a ranking of the relative importance of the input variables in explaining the performance of the fitted model on the test data.

Now, since we did not build our deep learning model in the Bayesian framework, the implementation of the procedure outlined above needs some adjustments. Specifically, we will replace the posterior distribution $\Pi(W\vert\D')$ by a point mass measure at $\hat W$, where $\hat W$ is the minimizer of (\ref{eq:loss}) obtained above from the training dataset $\D'$. The resulting approximation to the posterior distribution (\ref{post:dist}) is

\begin{equation}\label{profile:post:dist}
\Pi\left(\delta_1,\ldots,\delta_K\vert \D'\cup\D^{''}\right)\propto \exp\left(-r\sum_{j=1}^K \delta_j -\frac{1}{2\sigma^2}\mathsf{L}(\delta;\D)\right),
\end{equation}
where
\[\mathsf{L}(\delta;\D) = \sum_{t\in\D^{''}}\left\|\textbf{y}_t - \F_{\widehat{W}}(\delta\odot\bx_{t-h})\right\|_1.\]
We note that $\mathbf{L}$ depends both on the test set $\D{''}$ and the training set $\D'$.  We sample from (\ref{profile:post:dist}) using a straightforward Gibbs sampler. For an introduction to MCMC and the Gibbs sampler, we refer the reader to \citet{handbook:11}. Specifically, given $\delta\in\{0,1\}^K$, and $1\leq j\leq K$, let $\delta^{(j,1)}$ (resp. $\delta^{(j,0)}$) be the element of $\{0,1\}^K$, that is equal to $\delta$ except perhaps on component $j$, where $(\delta^{(j,1)})_j=1$ (resp. $(\delta^{(j,0)})_j=0$). Then under the joint (\ref{profile:post:dist}), the conditional distribution of $\delta_j$ given the remaining components denoted $\delta_{-j}$ and given $\mathcal{D}$ is a Bernoulli distribution $\textbf{Ber}(q_j)$, with probability $q_j$ given by
\[q_j = \left[ 1 + \exp\left\{r +\frac{1}{2\sigma^2}\left(\mathsf{L}(\delta^{(j,1)};\mathcal{D}) - \mathsf{L}(\delta^{(j,0)};\mathcal{D})\right)\right\}\right]^{-1}.\]
$q_j$ can be computed at the cost of two forward runs through the model $\mathcal{F}_{\hat W}$ using the test data sets with masks $\delta^{(j,0)}$ and $\delta^{(j,1)}$ applied.

\section{Results and Discussion}\label{sec:res}
\begin{figure}[ht]
\centering
\includegraphics[width=0.95\textwidth]{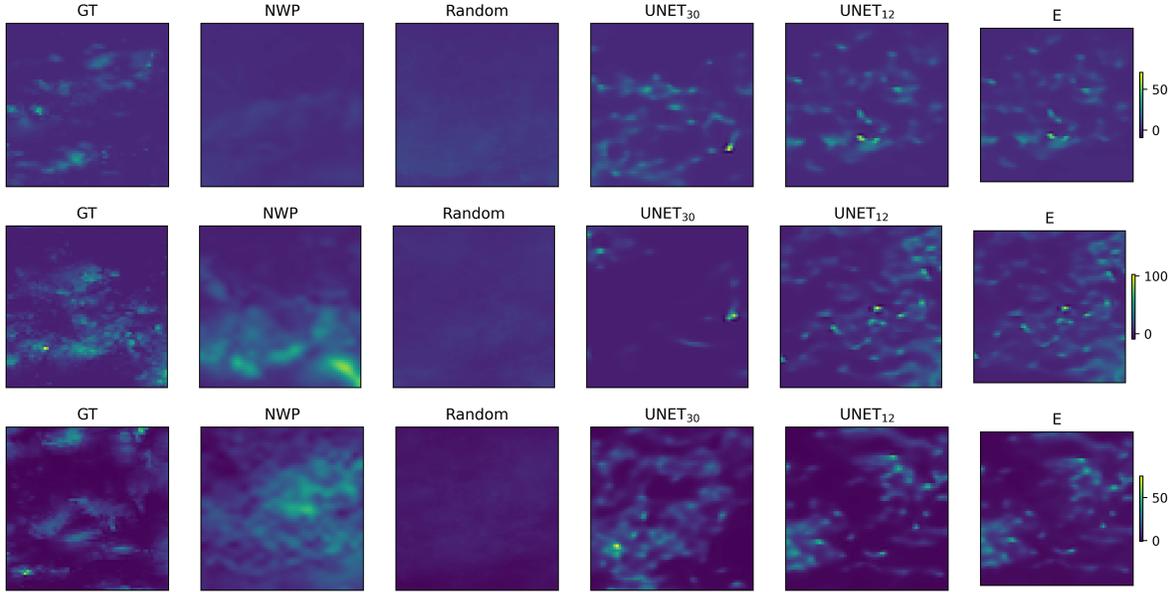}		\caption{Sample forecasts ($\rm{CLIM, NWP, UNET_{18}, UNET_{12}, HYB}$), along with the corresponding ground truth (GT) from GPM-IMERG.}
\label{fig:samples}
\end{figure}

We evaluate the performance of the models using the metrics presented above. As a start, Figure \ref{fig:samples} shows some random ground truth rainfall images, and their predictions from the models under comparison.

\subsection{Comparison in terms of mean absolute error and CRPS}
Table \ref{tab:mae} shows the overall mean and standard deviation of the MAE and CRPS computed across all pixels in the test dataset. The results show that in terms of the MAE, the $\rm{NWP}$ model is comparable to the climatology $\rm{CLIM}$, but $\rm{UNET_{12}}$, $\rm{UNET_{18}}$, and the HYB model show clear improvement. The fact that the numerical weather prediction model does not outperform the climatology forecast under the tropics is well-documented, as we discussed in the introduction \citep{Vogel2020}. However, climatology forecasts are broad averages and therefore lack precision, as we will see below.

\begin{figure}[h]
\centering
\caption{CRPS skill maps. We treat the spatial locations as independent. Each skill value is computed as in (\ref{eq:skill}). For increased sharpness of the figures, the skill values have been discretized into three categories: 1 (better than CLIM), 0 (equal to CLIM), and -1 (worse than CLIM). The UNET models show superior skill in most regions, while the NWP forecasts perform better in the southeast corner of the AOI.}
\includegraphics[width = 0.9\textwidth]{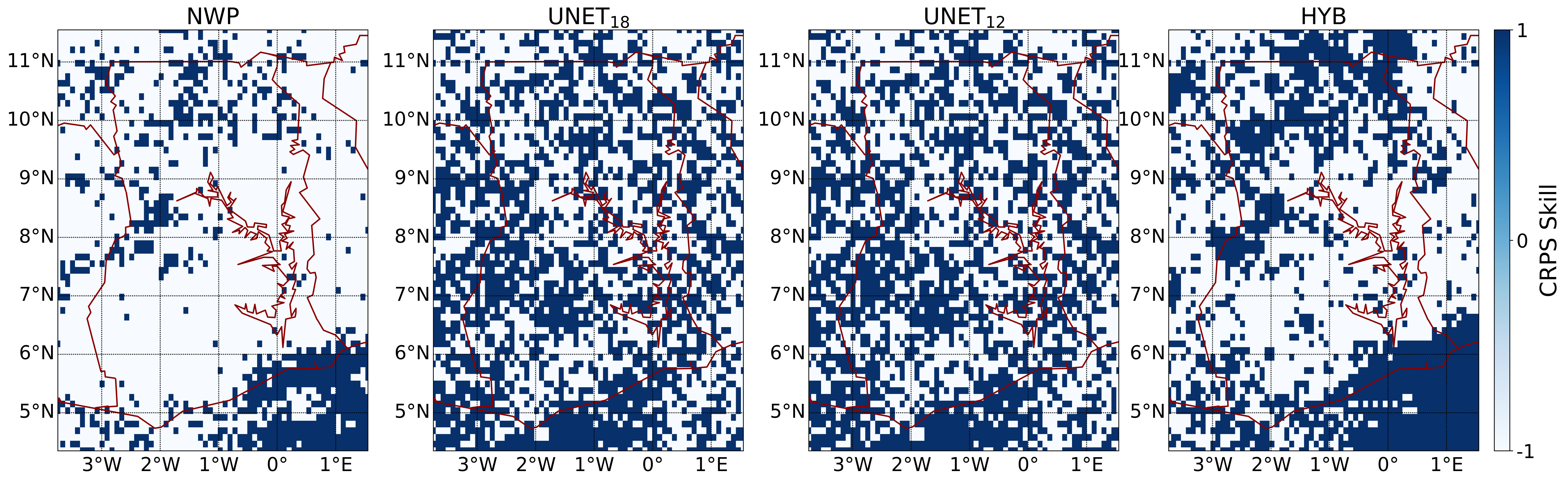}
\label{fig:mae:skills}
\end{figure}

Figure~\ref{fig:mae:crps} presents the location-wise CRPS maps for all comparative approaches. Overall, both UNET models perform better than $\rm{NWP}$ and $\rm{CLIM}$ across most regions, while the $\rm{NWP}$ model demonstrates stronger performance in the southeastern corner of the area of interest. Similarly, Figure \ref{fig:mae:skills} shows the CRPS skill maps of the models compared to  $\rm{CLIM}$ as given in (\ref{eq:skill}). White areas hold negative values and indicate locations at which the $\rm{CLIM}$ has better CRPS. The maps are consistent with the summary results on Table \ref{tab:mae} and show that the $\rm{CLIM}$ significantly under-performs $\rm{UNET_{12}}$ and $\rm{UNET_{18}}$. The hybrid method {\rm HYB} has improved MAE as shown on Table \ref{tab:mae} but has comparatively poor CRPS score. This is possibly due to our inability to apply EasyUQ to {\rm HYB} as we explain in section \ref{sec:method}-\ref{sec: easyuq}. Nevertheless, we note in Figure \ref{fig:mae:skills} that the hybrid model performs better than NWP and UNET12 in the northern and south-east regions, but worse in the central region.
\bigskip
\begin{table*}[ht]
\caption{Means and standard deviations (SD) of the Mean absolute errors (MAE) and Continuous Ranked Probability Score (CRPS)}
\label{tab:mae}
\centering
\begin{tabular}{|c|c|c|c|c|}
\hline
\textbf{Model} & \textbf{Means of MAE} & \textbf{SD of MAE} & \textbf{Means of CRPS} & \textbf{SD of CRPS} \\ \hline
CLIM & 3.90 & 1.15 & 2.98 & 1.27 \\ \hline
NWP & 3.92 & \textbf{1.00} & 3.94 & 1.16 \\ \hline
\textbf{$\rm{UNET_{18}}$} & 3.81 & 1.22 & \textbf{2.71} & \textbf{0.96} \\ \hline
\textbf{$\rm{UNET_{12}}$} & 3.74 & 1.13 & 2.72 & 0.96 \\ \hline
\textbf{HYB} & \textbf{3.69} & 1.01 & 3.01 & 0.80 \\ \hline
\end{tabular}
\end{table*}

\begin{figure}[h]
\centering
\caption{Location-wise CRPS map obtained using $\rm{CLIM, NWP, UNET_{18}, UNET_{12}, HYB}$ models}
\includegraphics[width = 0.9\textwidth]{CRPS.png}
\label{fig:mae:crps}
\end{figure}

\subsection{Comparing the biases of \rm{NWP} and $\rm{UNET_{18}}$} 
Here we further investigate the structure of the prediction biases of the models \rm{NWP} and $\rm{UNET_{18}}$. We define the bias as the difference between the predicted rainfall and the corresponding ground truth. Clearly, the closer the bias is to zero, the better. We combine all the pixels of all the test samples together, and look at the proportion of pixels with bias falling within each of the following ranges $(-\infty,-1], (-1,-0.1], (-0.1, 0]$, $(0,  0.1]$, $(0.1, 1]$, and $(1,\infty)$.  The results are presented in Table \ref{Dist}. We note that the proportion of bias values produced by $\rm{UNET_{18}}$ that fall in the combined interval $(-0.1,0.1)$ is $70\%$ higher than for \rm{NWP}. We also note that $\rm{UNET_{18}}$ tends to have more negative bias, compared to \rm{NWP} which is positively biased. This observation has initially motivated our construction of the hybrid method.

Although the individual pixel-wise forecasts are not independent, one can perform a homogeneity chi-square test (see e.g. \cite{agresti:2007}~Section 2.4), to check whether the two models have similar probabilities of landing their forecast biases in these 6 intervals. Let $p_{m,c}$ be the probability that at any given pixel, method $m\in\{{\rm NWP, UNET}\}$ produces a forecast with bias in interval $c\in \{1,\ldots,6\}$, where we use ${\rm UNET}$ as a shorthand for ${\rm UNET_{18}}$. Let $N_{m,c}$ be the number of times method $m\in\{{\rm NWP, UNET}\}$ produces a forecast with bias in interval $c$ as given in Table \ref{Dist}. Then, under the assumption that individual pixel-wise forecasts are independent, and under the null assumption that $p_{{\rm NWP},c} = p_{{\rm UNET},c}$ for $c=1,\ldots 6$, the statistic
\[S = \sum_{c=1}^6\frac{\left(N_{{\rm UNET},c} - N_{{\rm NWP},c}\right)^2}{\left(N_{{\rm UNET},c} + N_{{\rm NWP},c}\right)},\]
follows asymptotically a chi-square distribution with $5$ degrees of freedom. The value of $S$ is $S=39,426.78$, with p-value $= 0.0$ (up to the machine epsilon). The most salient difference between the two distributions is that ${\rm UNET}_{18}$ assigns higher probability to the combined interval $(-0.1,0.1]$. These results thus support the conclusion that the DL model $\rm{UNET_{18}}$ shows better performance compared to \rm{NWP}.

\bigskip
\begin{table}[h]
\centering
\caption{Counts and Proportions (in percentage) of pixels in each range}
\label{Dist}
\begin{tabular}{|l|cccccc|}
\hline
\multicolumn{1}{|c|}{\multirow{2}{*}{\textbf{Model}}} & \multicolumn{6}{c|}{\textbf{Counts and Proportions}}                                               \\ \cline{2-7} 
\multicolumn{1}{|c|}{}                                & \multicolumn{1}{c|}{\textbf{($-\infty$, -1]}} & \multicolumn{1}{c|}{\textbf{(-1, -0.1]}} & \multicolumn{1}{c|}{\textbf{(-0.1, 0]}} & \multicolumn{1}{c|}{\textbf{(0, 0.1]}} & \multicolumn{1}{c|}{\textbf{(0.1, 1]}} &
\multicolumn{1}{c|}{\textbf{(1, $\infty$)}} \\ \hline
$\rm{NWP}$                             & \multicolumn{1}{c|}{43237 (19.19)}                    & \multicolumn{1}{c|}{9546 (4.24)} 
& \multicolumn{1}{c|}{24404 (10.83)}
& \multicolumn{1}{c|}{32850 (14.58)}
& \multicolumn{1}{c|}{39088 (17.35)}
& 76155 (33.80)         \\ \hline
$\rm{UNET_{18}}$                       & \multicolumn{1}{c|}{47820 (21.23)}   
& \multicolumn{1}{c|}{18540 (8.23)}  
& \multicolumn{1}{c|}{17525 (7.78)} 
& \multicolumn{1}{c|}{79252 (35.18)}
& \multicolumn{1}{c|}{19671 (8.73)}
& 42472 (18.85)          \\ \hline
\end{tabular}
\end{table}

\subsection{Comparison in terms of precision and recall}
Correctly matching the tail events of the rainfall distribution has significant practical value. Here we compare our forecast models in their ability to correctly predict rainy days and heavy rain days (days with 24h rainfall larger than $0.5$mm, and $10$mm respectively). Using the GPM-IMERG as ground truth, we compare the models in terms of precision (P), recall (R), and F1-score as described in Section \ref{sec:eval}. Table~\ref{PRF1} summarizes the overall performance of each model.  We first consider the detection of rainy days (at threshold $\tau= 0.5$mm). As expected, due to broad averaging, the \rm{CLIM} forecast suffers from low precision ($0.45$), as it frequently misclassifies non-rain events as rain. Similarly, the $\rm{NWP}$ model also shows a strong recall ($88\%$) but a relatively low precision ($0.49$). This suggests that the $\rm{NWP}$ model captures much of the rainfall signal, but also has some level of spurious effects driving its forecasts. In contrast, the data-driven models have lower recall, which suggests that these models are still missing some of the important rainfall signal (the physics), however, they have better precisions. The hybrid model, $\mathrm{HYB}$, improves on  $\rm{NWP}$ with better recall and precision. The location-wise distribution of precision and recall across the area of interest is illustrated in Figure~\ref{fig:pr:05}.
\bigskip

\begin{table}[]
\centering
\caption{The overall Precision (P), Recall (R), and F1-Score (F1) of the models at the threshold, $\tau=0.5$ and $10$}
\label{PRF1}
\begin{tabular}{|l|ccc|ccc|}
\hline
\multirow{2}{*}{Model} & \multicolumn{3}{c|}{$\tau = 0.5$} & \multicolumn{3}{c|}{$\tau = 10$} \\ \cline{2-7} 
& \multicolumn{1}{c|}{P} & \multicolumn{1}{c|}{R} & F1 & \multicolumn{1}{c|}{P} & \multicolumn{1}{c|}{R} & F1 \\ \hline
CLIM & \multicolumn{1}{c|}{0.45} & \multicolumn{1}{c|}{\textbf{0.97}} & 0.61 & \multicolumn{1}{c|}{0.17} & \multicolumn{1}{c|}{0.02} & 0.04 \\ \hline
NWP & \multicolumn{1}{c|}{0.49} & \multicolumn{1}{c|}{0.88} & 0.62 & \multicolumn{1}{c|}{0.17} & \multicolumn{1}{c|}{0.08} & 0.11 \\ \hline
$\rm{UNET_{18}}$ & \multicolumn{1}{c|}{0.60} & \multicolumn{1}{c|}{0.61} & 0.61 & \multicolumn{1}{c|}{0.28} & \multicolumn{1}{c|}{0.21} & 0.24 \\ \hline
$\rm{UNET_{12}}$ & \multicolumn{1}{c|}{\textbf{0.62}} & \multicolumn{1}{c|}{0.64} & 0.63 & \multicolumn{1}{c|}{\textbf{0.31}} & \multicolumn{1}{c|}{\textbf{0.26}} & \textbf{0.28} \\ \hline
HYB & \multicolumn{1}{c|}{0.51} & \multicolumn{1}{c|}{0.90} & \textbf{0.65} & \multicolumn{1}{c|}{0.24} & \multicolumn{1}{c|}{0.12} & 0.16 \\ \hline
\end{tabular}
\end{table}

\bigskip

\begin{figure}[h]
\centering
\begin{tabular}{c}
\includegraphics[width = 0.9\textwidth]{Precision_05.png}\\
\includegraphics[width = 0.9\textwidth]{Recall_05.png}
\end{tabular}
\caption{First row: Locationwise recall (R) of the models at threshold $\tau= 0.5$mm. Second row: Locationwise precision (P) of the models at threshold $\tau= 0.5$mm.}
\label{fig:pr:05}
\end{figure}

We also compare the models through their predictions of heavy rainy days, particularly on how such predictions agree with GPM-IMERG. We set the threshold for heavy rain to $\tau =10$mm.  
The performance of the models is much worse. This suggests that more work is needed to better predict extreme rainfall events.  Figure~\ref{fig:pr:10} shows the location-wise distribution of P and R across the area of interest. From these figures and Table Table~\ref{PRF1}, we see again that $\rm{UNET_{12}}$ (and $\rm{UNET_{18}}$ to a lesser extent) performs better than $\rm{NWP}$. Here also the hybrid model, $\mathrm{HYB}$, has improved precision and recall compared to $\mathrm{NWP}$, but does worst than $\mathrm{UNET_{18}}$.

\bigskip

\begin{figure}[h]
\centering
\begin{tabular}{c}
\includegraphics[width = 0.9\textwidth]{Precision_10.png}\\
\includegraphics[width = 0.9\textwidth]{Recall_10.png}
\end{tabular}
\caption{First row: locationwise precision (P) of the models at threshold $\tau= 10$mm. Second row: locationwise recall (R) of the models at threshold $\tau= 10$mm.}
\label{fig:pr:10}
\end{figure}

\subsection{Input variables selection}
\begin{figure}[h]
\centering
\includegraphics[scale = 0.17]{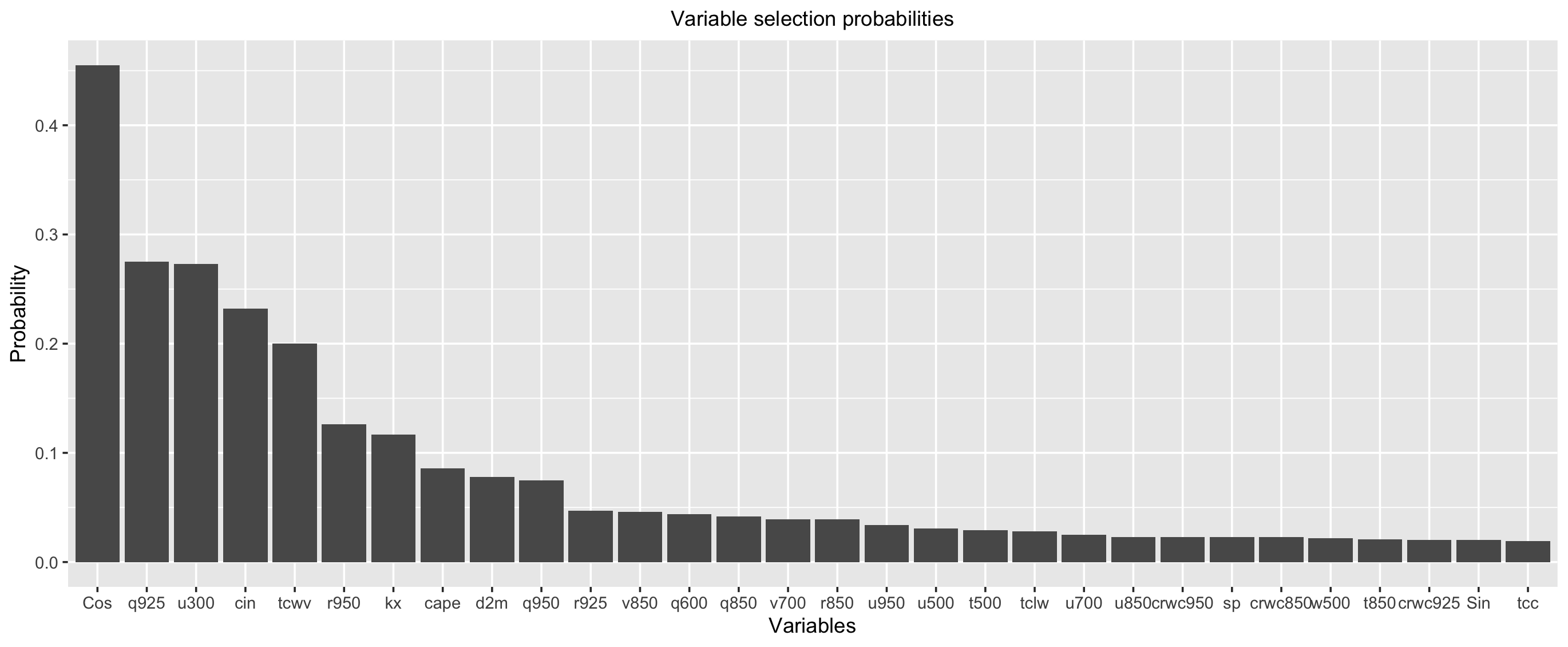}
\caption{Top 30 marginal posterior means of input variables using samples from the posterior distribution (\ref{profile:post:dist}).}
\label{fig:var:sel}
\end{figure}

We run the Gibbs sampler described in Section \ref{sec:varsel} for $1,000$ epochs, where each iteration is a full sweep through the $K$ variables. We use $\sigma^2=0.01$, and $r=3.76$. We discard the first 950 epochs as burn-in. For each input variable $j$, we take the average of $\delta_j$ along the MCMC output as an estimate of the posterior probability of $\delta_j$. Figure \ref{fig:var:sel} shows the $30$ largest posterior probabilities. The results show that the most important predictive variable is the spatio-temporal variable $\rm{COS}$ defined in (\ref{def:cos}). This is hardly surprising since rainfall in Ghana is known to have a strong seasonality which varies with latitude. Evaporation drives rainfall, and our method indeed highlights  specific humidity ($q925$),  relative humidity ($r950$), and total column water vapor ($tcwv$) as important input variables. The variable wind ($u300$) also appears important. The wind variable $u300$ is possibly related to the Tropical Easterly Jet (TEJ), which plays a fundamental role in the West African monsoon \citep{nicholson2013west}]. 
Our methodology also highlights several convection-related variables: convective inhibition ($cin$), K-index ($kx$) and the convective available potential energy ($cape$) as key input variables. This is realistic, since a large part of the rainfall in Ghana is convection-driven. One important limitation of our methodology is that it does not offer much insight on the physical process(es) by which these variables affect rainfall. More research is needed on this point.

\section{Conclusion}\label{sec:conclusion}
This research dealt with the challenging problem of predicting rainfall in tropical regions. Although we have focused on Ghana, our methodology can be readily applied to other regions. Our proposed data-driven model is a U-Net convolutional neural network model (CNN) that we trained to predict 24h rainfall at 12h and 18h lead-time using data from ERA5 and GPM-IMERG. We find that the 12h lead-time prediction model shows improvement in terms of precision and accuracy over a numerical weather prediction (NWP) model developed by the ECMWF as obtained through TIGGE. We also explore with mixed results a hybrid approach that combines the U-Net model with the ECMWF forecasts. In particular, while the hybrid model achieves the lowest MAE among all models and generally outperforms the NWP, it performs worst than the stand-alone U-Net models in terms of CRPS and the ability to detect large rainfall events. These results suggest that more sophisticated methods are needed to fully leverage the potential of the hybrid approach. With the dazzling rise of deep learning in weather forecasting, this issue stands as an important direction for future research. 
 Finally, we also developed a novel methodology that enhanced the interpretability of our model by identifying the input variables that best capture the performance of the model. 

Our work comes with several limitations, but also opens up several avenues for further research that we hope to tackle in the future. Although we have identified a small list of variables that appear to play an important role in the forecasts, the precise physical processes by which these variables impact rainfall remain largely unclear.  There is a growing literature that combines statistics, machine learning, and partial differential equations and aims to discover new physical laws from data \citep{raissi:etal:19,Chen2020Physicsinformed}. Extending these methods to atmospheric sciences can provide a powerful empirical framework for identifying the physical processes involved in the present study.

Another notable shortcoming of our method is its poor performance in predicting heavy and extreme rainfall events, driven mainly by the lack of sufficient tail event data.  Models for better prediction of tail-events in statistics are often built using quantile regression. However, quantile regression at extreme quantile levels suffer from the same lack of tail data as in this study. New methodologies are thus needed to address this issue, perhaps along the emerging literature of extreme quantile regression that merges classical quantile regression and extreme value theory \citep{volthoen:2023,pasche:2024}. Expanding the area of interest and/or turning to gauge datasets such as the precipitation products of the Global Precipitation Climatology Centre \citep{gpcc:2014} may also prove useful in addressing this issue.

Finally, we acknowledge that, due to the current 5-days latency on the ERA5 dataset, our forecast models are currently not operational. This delay in data availability, (which affects all models, including foundation models built on ERA5) clearly creates a barrier to the widespread deployment of data-driven weather forecasting. 

\section*{Acknowledgments}
We are grateful to Nils Cazemier for assistance with the GPM-IMERG data product. We are grateful to Tilmann Gneiting for helpful discussions, and for pointing out several relevant papers. We also acknowledge the financial support of the National Science Foundation grant DMS-2210664. The authors also declare no conflicts of interest related to this study.

\section*{Data and Code Availability}
This research was conducted entirely from publicly available data for research purpose. The code to process the data and train our models are available at the Github address \texttt{https://github.com/indrakalita/RainfallForecasting}. The address also includes a script that utilizes the cdsapi interface to download the ERA5 data. We have used the website interface \texttt{https://gpm.nasa.gov/data/directory} and (\texttt{https://apps.ecmwf.int/datasets/data/tigge/levtype=sfc/type=cf/} to download the GPM-IMERG and TIGGE data, respectively.

\bibliographystyle{unsrtnat} 
\bibliography{RainfallForecasting_Arxiv.bib}

\end{document}